\documentclass[runningheads]{llncs}
\usepackage[T1]{fontenc}
\usepackage{graphicx}
\usepackage{url}
\usepackage{booktabs}

\usepackage{tabularx}
\usepackage{multirow}
\begin{document}

\title{Distantly Supervised Morpho-Syntactic Model for Relation Extraction}
\author{Nicolas Gutehrlé\inst{1}\orcidID{0009-0000-1958-215X} \and
Iana Atanassova\inst{1,2}\orcidID{1111-2222-3333-4444}
}

% \author{Anonymous authors}
%
% \authorrunning{Gutehrlé et al.}
\institute{Université de Franche-Comté, CRIT \\
            F-25000 Besançon, France\\
\and
Institut Universitaire de France (IUF)
% \email{lncs@springer.com}
}
\maketitle              % typeset the header of the contribution
\begin{abstract}
     The task of Information Extraction (IE) involves automatically converting unstructured textual content into structured data. Most research in this field concentrates on extracting all facts or a specific set of relationships from documents. In this paper, we present a method for the extraction and categorisation of an unrestricted set of relationships from text. Our method relies on morpho-syntactic extraction patterns obtained by a distant supervision method, and creates Syntactic and Semantic Indices to extract and classify candidate graphs. We evaluate our approach on six datasets built on Wikidata and Wikipedia. The evaluation shows that our approach can achieve Precision scores of up to 0.85, but with lower Recall and F1 scores. Our approach allows to quickly create rule-based systems for Information Extraction and to build annotated datasets to train machine-learning and deep-learning based classifiers. 

     %Most errors can be corrected by increasing the variety of morpho-syntactic extraction patterns stored in the Syntactic Index. 


\end{abstract}

\section{Introduction}
\label{sec:introduction}

The Information Extraction (IE) task consists in automatically transforming unstructured textual content into structured data \cite{jurafsky-martin-speech-2008}, usually as \textit{(subject, relation, object)} triples. These triples can then be used for multiple downstream tasks such as filling a relational database, building a Knowledge Base (KB) or applying logical reasoning to infer new facts. The IE task can be divided into two main categories: Open and Closed Information Extraction.

On the one hand, Open Information Extraction (OIE) systems extracts as many facts as possible from documents. Such systems do not require annotated datasets to be built. However, they tend to extract facts that are either too general or too specific, thus limiting their usefulness. On the other hand, Closed Information Extraction (CIE) systems extract specific types of facts from the documents. It is usually cast as a Relation Extraction (RE) task, which consists in identifying the relation between two Named Entities. Although they usually perform better than OIE systems, CIE systems require annotated datasets, which are costly to produce. Moreover, casting this task as Relation Extraction usually requires to pre-process the documents with Named Entity Recognition (NER). Similarly, training a NER classifier requires a dataset of labelled examples, which is difficult to produce, especially for domain-specific entities, as in the medical or biology domains. 

In this paper, we present a method for the extraction and categorisation of an unrestricted set of relationships from text. Our method relies on a Syntactic Index of high Precision morpho-syntactic extraction patterns expressing specific relations or facts about entities. Those patterns are collected from Wikidata and Wikipedia in a distant supervision manner \cite{mintz-etal-2009-distant}, which is often used to build weakly annotated Relation Extraction datasets. Moreover, we describe a method to train a Semantic Index inspired by Explicit Semantic Analysis \cite{Gabrilovich2007ComputingSR} which is used to classify candidate graphs. Our method allows to collect extraction patterns for any relation described in the Wikidata ontology in any language available on Wikidata and Wikipedia. Moreover, our method only requires to pre-process documents with POS-tagging and dependency parsing, using tools that are freely available.

The rest of the paper is structured as follows: Section \ref{sec:relatedwork} presents related works on Information Extraction. We describe our method to collect and build our dataset from Wikipedia and Wikidata in Section \ref{sec:dataset}. In Section \ref{sec:methodology}, we describe our method to build the Syntactic and Semantic Indices.  Finally, we evaluate our methodology in Section \ref{sec:evaluation} and present our conclusions and suggestions for future works in Section \ref{sec:conclusion}.

% most work train classifiers which either rely on machine or deep learning architectures
% we propose a method which relies on morpho-syntactic patterns to extract relations and facts from text, and simple features such as TF-IDFfor classification

% Our method relies on morpho-syntactic extraction patterns obtained from Wikidata and Wikipedia by a distant supervision method \cite{mintz-etal-2009-distant}, and creates Syntactic and Semantic Indices to extract and classify candidate graphs. Those patterns can be used for Relation Extraction or even to extract facts about an entity.

% by either applying hand-crafted extraction patterns or by training a classifier which learns such patterns in a self-supervised manner.
% These methods can process large amount of data as in the World Wide Web, and should be domain-independent.

% Moreover, a further step of mapping the triples to a predefined set of categories is needed in order to use them for downstream tasks such as Knowledge Base population. 

% In general, these systems reach better results than Open systems. However, they require labelled datasets of the types of relation to identify. 

% data from Wikipedia and Wikidata and build a weakly annotated dataset of dependency graphs expressing specific relations. 

% We describe our method for extracting information about an entity in Section \ref{sec:methodology}. In Section \ref{sec:dataset}, we present the implementation of our methodology and the datasets we have created for our experiment.

\section{Related Work}
\label{sec:relatedwork}

\subsection{Open Information Extraction}

Open Information Extraction (OIE) \cite{yates-etal-2007-textrunner} aims at extracting all possible facts from documents in an unsupervised manner. OIE systems are domain-independent, and must be able to extract facts regardless of the type of document. The extracted facts are represented as $(arg1;relation;arg2)$ triples. OIE systems are usually rule-based, self-supervised or unsupervised. Rule-based methods \cite{akbik-loser-2012-kraken,fader2011identifying,white2016universal} extract facts by applying hand-crafted extraction patterns based on linguistic features such as part-of-speech or dependency trees. Self-supervised methods \cite{mausam-etal-2012-open,wu-weld-2010-open,yahya-etal-2014-renoun,yates-etal-2007-textrunner} train a classifier on a set of patterns that have been automatically extracted from an unlabelled dataset. Recent works rely on neural network architectures \cite{zhou2022survey} and cast this problem as a sequence-labelling task \cite{stanovsky-etal-2018-supervised} or a sequence-generation task \cite{cui2018neural}. Some systems, e.g. \cite{bhutani-etal-2016-nested,del2013clausie}, pre-process sentences to extract clauses from sentences, thus improving accuracy. 

Although OIE systems can extract a vast amount of facts and are scalable to large amount of data such as the World Wide Web, they tend to extract facts that are either too general or too specific, thus limiting their usefulness \cite{pawar2017relation}. Moreover, most methods are language-dependent: most available systems such as \cite{fader2011identifying} are tuned to English. Only a few works such as \cite{zhila2014open} are applied to other languages, whereas works such as \cite{white2016universal} are language-agnostic or multilingual \cite{ro-etal-2020-multi}. Because of the amount of extracted facts, most works evaluate their system in terms of Precision and ignore the Recall measure. Since OIE systems must be domain-independent, building a dataset to evaluate them is a difficult task \cite{niklaus-etal-2018-survey}. Thus, most work evaluate their methods by examining a small sample of their results, although recent works by \cite{schneider-etal-2017-analysing,stanovsky-dagan-2016-creating} have proposed benchmarks to evaluate OIE systems.

Unsupervised relation extraction, sometimes called Relation Discovery (RD) \cite{hasegawa-etal-2004-discovering}, is another form of OIE, where clustering methods identify sets of patterns expressing the same concept from an unlabelled dataset. First, entity pairs and their contexts are extracted and transformed into features, which are usually morphological, lexical and semantic, before being clustered by similarity \cite{alfonseca-etal-2012-pattern,hasegawa-etal-2004-discovering,Lin2001DIRTD,yan-etal-2009-unsupervised,yao-etal-2011-structured,yao-etal-2012-unsupervised}. Some works add constraints such as entity types constraint \cite{hasegawa-etal-2004-discovering,yao-etal-2011-structured} or external knowledge \cite{lopez-de-lacalle-lapata-2013-unsupervised} in order to orient the clustering step. Similarly, recent works \cite{ali2021unsupervised,Elsahar_2017} relied on word embeddings or language models such as BERT \cite{devlin-etal-2019-bert} to encode sentences into dense-vectors, then cluster them by similarity.

Many works \cite{ali2021unsupervised,Elsahar_2017,lopez-de-lacalle-lapata-2013-unsupervised,yan-etal-2009-unsupervised,yao-etal-2011-structured} have built their methods on labelled datasets such as the New York Time dataset \cite{Riedel2010ModelingRA} or Knowledge Bases such as Freebase or Wikidata, which are usually used for Closed Information Extraction methods. Thus, most unsupervised systems are evaluated in terms of Precision, Recall and F1-scores, where they compare the discovered relations with those from the datasets. Since they don't rely on labelled datasets, unsupervised relation classifiers are quick to build and are able to identify new relations from the datasets. However, they are often less precise than supervised methods. Moreover, a further step to label the clusters is necessary.

\subsection{Closed Information Extraction}

Closed Information Extraction (CIE), usually referred as Relation Extraction (RE), aims at extracting a specified set of relations between entities. Those systems go from rule-based to supervised. Rule-based systems such as \cite{10.1145/2611040.2611079,hearst-1992-automatic,nebhi2013rule} detect and classify the relations between entities by applying manually or semi-automatically crafted rules. These systems usually have high Precision scores but lower Recall scores, and require experts to conceive them. Bootstrapping methods \cite{Brin1998ExtractingPA} allow to accelerate the conception of extraction patterns. Given a set of initial seeds of entity pairs involved in a given relation, a first set of patterns is extracted from an unlabelled dataset. These new patterns are then used to find new pairs of entities. This process is repeated until no new patterns of entity pairs are found. Works such as \cite{10.1145/336597.336644,phi-etal-2018-ranking} evaluate the extracted patterns at each iteration and add constraints, such as entity types, in order to only keep the best patterns. The bootstrapping method allows to rapidly extract patterns expressing a relation if the initial seeds are of good quality. However, the Precision scores of the extracted patterns tend to be low. They also often suffer from semantic drift, i.e. the relation expressed in the extracted patterns diverge from the relation initially searched.

Most supervised methods are machine-learning based, where a classifier is trained on a labelled dataset. They are generally divided into kernel-based and feature-based methods. Kernel-based methods \cite{bunescu-mooney-2005-shortest,10.5555/2976248.2976270,collins2001convolution,csse.2022.030759} use a kernel function to compute the similarity between samples and classify them with a Support Vector Machine (SVM) whereas feature-based \cite{chan-roth-2011-exploiting,kambhatla-2004-combining,nguyen-etal-2007-subtree,rink-harabagiu-2010-utd} extract features from the samples to train a classifier. Recent works use word embeddings or language models to represents texts as feature vectors and train neural-network based models \cite{https://doi.org/10.48550/arxiv.1810.10147,wang2022deep,zhang-etal-2018-graph}. Supervised methods obtain good results but they require  labelled datasets, which are not always available and are costly to produce. 

The distant supervision method \cite{mintz-etal-2009-distant} is a possible solution to the issue of the lack of dataset. It assumes that if two entities participate in a relation in a Knowledge Base, e.g. Freebase or Wikidata, then at least one sentence which mentions both entities in a document  expresses that relation \cite{Riedel2010ModelingRA}. Thus, this method allows to quickly build a dataset of weakly annotated samples. However, the risk to annotate a sample with the wrong label is high, since the same set of entities may participate in several different relations. 

Most works focus on extraction relations at the sentence level, i.e. between entities in the same sentence. Recent works \cite{christopoulou-etal-2018-walk,quirk-poon-2017-distant,swampillai-stevenson-2011-extracting,zeng-etal-2017-incorporating,zhou2020documentlevel} focus on document-level RE and aim at also extracting inter-sentence relations, i.e. relations between entities in different sentences. Document-level RE is however much harder than sentence-level RE since the same entity pair may participate in different relations and the contextual information needed to classify the relation is harder to find. It also requires more pre-processing steps such as co-reference resolution or discourse analysis \cite{han2020data}. Finally, some recent works \cite{Bekoulis_2018,chen-etal-2020-joint-entity,popovic-etal-2022-aifb} tackle the Named Entity Recognition (NER) and RE tasks jointly, instead of sequentially. They presume that the presence of two entities in a sentence suggests the presence of a relation, and vice-versa. Although such methods yield good results, the amount of candidate spans that need to be verified is computationally expensive.

Closed Information Extraction systems are usually evaluated in terms of Precision, Recall and F1 scores. The Relation Extraction task has attracted a lot of interest in the recent years. Thus, many datasets have been created to conceive and evaluate these systems. Most available datasets such as CoNLL2004 \cite{roth-yih-2004-linear}, ACE \cite{doddington-etal-2004-automatic}, Sem-Eval \cite{girju-etal-2007-semeval,hendrickx-etal-2010-semeval}, the New York Times (NYT) dataset \cite{Riedel2010ModelingRA} or TACRED \cite{zhang-etal-2017-position} have been created on news articles and web pages. Others are domain-specific, such as the ScienceIE \cite{augenstein-etal-2017-semeval}, SciERC \cite{luan-etal-2018-multi} or SemEval2018 \cite{gabor-etal-2018-semeval} datasets which were made on scientific documents. Others have been created on Wikipedia such as the FewRel \cite{han-etal-2018-fewrel}, mLAMA \cite{kassner-etal-2021-multilingual} or SMiLER \cite{seganti-etal-2021-multilingual}. Most of these datasets are monolingual and contain examples in English. There are however some multilingual datasets such as mLAMA, or SMiLER. Moreover, most datasets focus on Relation Extraction at the sentence level. There are however a few datasets such as DocRED \cite{yao-etal-2019-docred} and DWIE \cite{zaporojets2021dwie} which also focus on RE at the document level.  

\section{Dataset}
\label{sec:dataset}

% In this section, we present our methodology for extracting and classifying information about an entity. It consists in several main steps: first, we collect a dataset of annotated dependency graphs in a distant supervision manner from Items of a given type in Wikidata. 

% In this section, we describe Wikidata, the most comprehensive Knowledge Base available today, which we use to build our weakly annotated dataset of dependency graphs.

\begin{figure}[htbp]
    \centering
    \includegraphics[width=.8\textwidth]{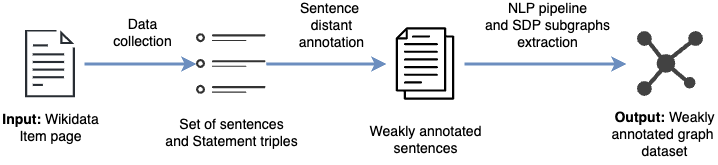}
    \caption{Pipeline for collecting and processing data}
    \label{fig:wikidatapipeline}
\end{figure}

In this section, we describe our pipeline to extract data from Items in Wikidata and  build a weakly annotated dataset of dependency graphs expressing specific relations between entities. We describe how data in Wikidata are structured, and then we present in details our data collection pipeline, which is shown in Figure \ref{fig:wikidatapipeline}. Finally, we describe the datasets that we have built by applying this pipeline.

\subsection{Description of Wikidata}
\label{sec:wikidata}

Wikidata is made of Items, which describe anything from a person, a place to a concept. Items are given a label, a description and a set of aliases. For instance in English, the Item Q23 has the label "George Washington", the description "president of the United States from 1789 to 1797" and the aliases "Father of the United States", "The American Fabius" and "American Fabius". Labels, descriptions and aliases are usually available in multiple languages. An Item also stores links to related Wikimedia pages, such as Wikipedia, Wikiquote or Wikinews.
% Each Item is given an id starting with the letter Q.

Information about an Item are recorded as \textit{(Source;Property;Target)} triples named Statements, such as \textit{(George Washington, date of birth, 22 February 1732)} or \textit{(George Washington, place of burial, Mount Vernon)}. The \textit{Property} element in the Statement triple are special Items called Properties, which describe the relationship between Items. Similarly, each Property has a label, a description and possible aliases. For instance in English, the Property P19 has the label "place of birth". The \textit{Source} element of the Statement is the current Item $I$, for instance the Item Q23. The \textit{Target} element of a Statement is usually another Item, as is the case for the "place of birth" or "educated at" Properties. It may be a date as is the case for the "date of birth" or "date of death" Properties. It may also be a quantitative or textual value, which is the case for Properties such as "number of children" or "height". Finally, it may be empty. A Statement may record multiple values for the same Property.

\subsection{Data processing}
\label{sec:dataprocessing}

The first main step of our pipeline consists in collecting data related to an Item $I$. First, we collect the designations $id$ of $I$, i.e its label and aliases. Next, we extract the body of the Wikipedia page associated with $I$. For our experiment, we processed each Item to collect data written in French. The content of related Wikipedia pages was collected by extracting the content of <p> tags from their HTML pages. This content was then divided into sentences using simple heuristic rules written as regular expressions. In order to ensure good linguistic analysis later, we clean these sentences with regular expressions to remove brackets and their content (often phonetic transcriptions), sentences starting with non-alphanumeric characters, trailing spaces and isolated non alphanumeric symbols.

Then, we extract the set $ST$ of non-empty Statements triples $st$ associated with $I$. We may extract all Statements or a selected subset. For our experiment, we chose to collect extraction patterns for the following properties: \textit{place of birth, date of birth, date of death, occupation, spouse, educated at}. Any unlabelled sentence is labelled as "\textit{Other}", so as to ensure the dataset contains negative samples. For such sentences, we ignore the SDP subgraph extraction step described further below. For each selected Statement triple $st$, we extract the designations $td$ of the \textit{Target} $T$. If $T$ is another Item, we extract its designations as we have just done for $I$. If $T$ is a date, we automatically translate it to the language that we need. For instance, "22 February 1732" in English would be translated into "22 février 1732" if we are working on French. Otherwise, if the value of $T$ is numerical or textual, we take it as is. 

As a result, each selected Statement $st$ is represented as a triple $(id, p, td)$, where $id$ are the designations of $I$, $p$ is the Property and $td$ are the designations of $T$. For instance, the Statement for the Property P509 ("cause of death" in English) for the item Q23 would give the triple \textit{("George Washington|Father of the United States|The American Fabius|American Fabius", P509, "epiglottitis|acute laryngitis")}. This first step outputs the set $SE$ of sentences $se$ from the related Wikipedia page and the set $ST$ of selected Statement triples $st$.

The second main step of the pipeline consists in annotating $SE$ in a distant supervision manner. For each Statement triple $st$, we search for the presence of $id$ and $td$ in every sentence $se$. To do so, we apply fuzzy matching to avoid missing a designation because of small orthographic differences such as a plural form. Thus, $se$ is labelled with the property $p$ of $st$ and associated with $st$ if it contains two sequences of characters that are respectively at least $N$\% similar to $id$ and $td$.  For instance, any sentence containing a designation of the preceding triple example is labelled P509, i.e. "cause of death", and becomes associated with this triple. For our experiment, we set $N= 90\%$. The same sentence may be labelled with multiple properties. This step outputs the set $WSE$ of weakly labelled sentences $wse$, where each sentence is associated with at least one Statement triple. 

\begin{figure}[htbp]
    \centering
    \includegraphics[width=.7\textwidth]{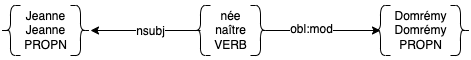}
    \caption{SDP subgraph between "Jeanne" and "Domrémy" for the sentence "Jeanne d'Arc est née à Domrémy" (\textit{Joan of Arc was born in Domrémy}). Each node is represented by its text, its lemma and its part-of-speech. The node labelled "née" (\textit{born}) is the anchor node.}
    \label{fig:graphexample}
\end{figure}

As assumed by \cite{bunescu-mooney-2005-shortest}, the relation between two entities in a sentence is expressed in the Shortest Dependency Path (SDP) between them. Thus, the third main step of the pipeline consists in processing the weakly labelled sentences $wse$ of $WSE$ with NLP tools so as to extract the SDP between $id$ and $td$ for each associated Statement triple $st$, and generate a weakly annotated dataset of dependency graphs. First, we apply a standard NLP pipeline to transform $wse$ into a directed dependency graph, where nodes are tokens and edges are the dependency relationship between two tokens. For our experiment, we processed these sentences with the French transformer model $fr\_core\_news\_trf$ provided by the \textit{spaCy} NLP framework to obtain the POS tag of words and dependency tree of the sentences. We store the textual form, the lemma and the part-of-speech of a node in its metadata. Secondly, we identify the graph's anchor node, i.e. the node with only out degrees and no in degree. Then, for each Statement $st$ associated with $wse$, we extract the SDP between $id$ and $td$ from the dependency graph. Since our sentences are not pre-processed with NER, we must first find the Source and Target node, i.e. the token respectively corresponding to $id$ and $td$. To do so, we apply fuzzy matching to find the node that is at least $N$\% similar to $id$ and to $td$. Again, we set $N = 90\%$ for our experiment. If a designation is made of multiple tokens, we select the root node of the matched group as the Source or Target node. From there, we can extract the SDP subgraph between the Source and Target nodes from the sentence's dependency graph. An example of a SDP subgraph is shown in Figure \ref{fig:graphexample}. The syntactic analysis may yield incorrect results and prevent from extracting the SDP subgraph. If so, the graph is discarded. This step outputs a set $WG$ of weakly labelled SDP graphs $wg$. 

\subsection{Data Distribution}
\label{sec:datadistribution}

We have created a weakly annotated dataset of dependency graphs by collecting data from the first 5000 Items of type Q5 ("human") returned by the "What Links Here" section of Wikidata\footnote{\url{https://www.wikidata.org/w/index.php?title=Special:WhatLinksHere/}}, which lists Items by type. We only select Items which are associated with a Wikipedia page in French. However, this method could work with any language available on Wikidata and Wikipedia. This dataset was then divided into a train, development and test set. In order to evaluate the impact of distant supervision over our method, we have also created ground truth (GT) train, development and test sets. These ground truth datasets have been manually created by selecting a subset of correctly labelled sentences from the weakly annotated datasets. The distribution of labels in the six datasets is shown in Table \ref{tab:labeldistrib}. Since we are building the Syntactic and Semantic Indices to recognise specific properties, there are no graphs labelled "Other" in both train sets. 

\small
\begin{table}[htpb]
    \centering
    \caption{Label distribution in the four datasets}

    \begin{tabular}{l|rrr|rrr}
    \hline 
    & \multicolumn{3}{c|}{Weak} & \multicolumn{3}{c}{GT} \\
    {}           &  Train   & Dev   & Test  & Train     & Dev   & Test \\
    \hline
    spouse       &  386     &  92   &  67   &   28      &   6   &  14 \\
    dateOfBirth  &  373     &  77   &  73   &   88      &  21   &  27 \\
    educatedAt   &  367     &  76   &  81   &   20      &   2   &   3 \\
    dateOfDeath  &  355     &  68   &  72   &   60      &  17   &   8 \\
    placeOfBirth &  353     &  70   &  77   &  117      &  21   &  16 \\
    occupation   &  348     &  78   &  80   &  266      &  57   &  54 \\
    Other        &  0       &  83   &  95   &   0       &  62   &  64 \\
    \hline

    \textbf{Total}        &  2182    & 544   &  545  &   579     &  186  &  186 \\
    \hline
    
    \end{tabular}
    \label{tab:labeldistrib}
\end{table}
\normalsize

\section{Method}
\label{sec:methodology}

In this section we present the Syntactic and Semantic Indices. First, we describe what those indices are, before explaining in details how to build them. Finally, we describe a pipeline which combines these indices to extract and classify facts from sentences.

\subsection{Building the Syntactic and Semantic Indices}
\label{sec:indexes}

The Syntactic Index stores morpho-syntactic extraction patterns which express one or multiple relations. These patterns are indexed by their anchor node, which were identified during the data processing step (see Section \ref{sec:dataprocessing}). Moreover, each graph is associated with a set of possible labels, i.e. the set of properties it has been labelled with during the data processing step. The Semantic Index is a  $W \times P$ matrix where $W$ is the number of unique words $w$ and $P$ is the number of unique properties $p$. The association between $w$ and $p$ is measured as a TF-IDF weight. While this approach is similar to the weighted inverted index used for Explicit Semantic Analysis (ESA) \cite{Gabrilovich2007ComputingSR}, our Semantic Index is not built on whole Wikipedia pages but on the textual content of SDP subgraphs. Moreover, whereas the concepts in the ESA index come from the title of the Wikipedia pages, ours come from the Property with which a graph has been labelled as concept.

\begin{figure}[htbp]
    \centering
    \includegraphics[width=.5\textwidth]{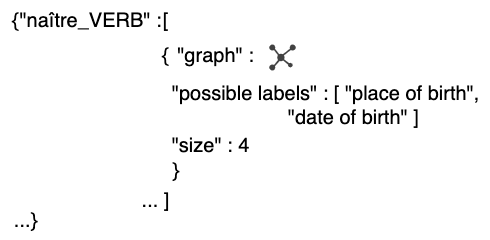}
    \caption{A graph in the "naître\_VERB" (\textit{born}) entry in the Syntactic Index}
    \label{fig:syntacticindex}
\end{figure}

To build the Syntactic Index $SynInd$, we first create an entry for each anchor node in our dataset, then add to each entry the corresponding graphs. In a second step, we remove duplicate graphs from each entry. A candidate graph $CG$ is considered a duplicate of a graph $G$ if they are isomorphic. Since our graphs are labelled, directed and acyclic, they must meet the following conditions to be isomorphic: they must have the same size, i.e. the same amount of edges, their edge connectivity must be identical and the nodes and the edges must have the same labels. These conditions allow us to ensure the graphs are syntactically and morphologically identical. If $CG$ and $G$ are isomorphic, we add the possible labels $pl$ of $CG$ to the set of labels of $G$. $CG$ is then removed from the anchor's list of graphs. Moreover, we store in the metadata of $G$ if it is ambiguous, i.e. if it has more than one possible label, and its size. The operation is repeated until every duplicate graph has been removed. For our experiment, we have built two Syntactic Indices, one on the Weak Train set and another on the GT Train set. The anchor nodes are represented as a concatenation of their textual value and their part-of-speech tag, as shown in the example on Figure \ref{fig:syntacticindex}. The statistics of both indices are shown in Table \ref{tab:synindexstatistic}. 
%If $CG$ and $G$ are isomorphic, we add the possible labels $pl$ of $CG$, i.e. the labels associated with $CG$, to the possible labels of $G$. 

\begin{table}[htpb]
    \centering
    \caption{Anchors and patterns distribution in the Weak and GT Syntactic Indices}
    \begin{tabular}{lrr}
    \hline
                            &  Weak Syntactic Index &  \ \  GT Syntactic Index   \\
            \hline
         Unique anchors                     &   497             &   173                   \\
         Unique patterns                    &   1024            &   335                 \\
         Ambiguous patterns                 &   3               &   0                   \\
         Mean number of pattern by anchor   &   2               &   2       \\
         \hline
    \end{tabular}
    \label{tab:synindexstatistic}
\end{table}

In order to build the Semantic Index $SemInd$, we first produce a $P \times W$ word frequency matrix, where $P$ is the number of unique properties $p$ and $W$ is the number of unique words $w$. To produce this matrix, we measure for each $p$ the frequency of each word $w$ occurring in each document $d$ labelled with $p$. Secondly, we transform this matrix into a TF-IDF matrix so as to obtain an association weight between words and properties. This matrix is finally transposed into a $W \times P$ matrix to obtain the Semantic Index. An example of a Semantic Index is given in Table \ref{tab:semanticindex}. Similarly, for this experiment, we have built two Semantic Indices, one on the Weak Train set and another on the GT Train set. The statistics of both indices are shown in Table \ref{tab:semindexstatistic}.

% In order to build the Semantic Index $SemInd$, we first produce a $D$ x $W$ where $D$ is the number of documents $d$, i.e. the SDP graphs, and $W$ is the number of unique words $w$ in the dataset.  We select every word in the dataset excepted proper nouns. This matrix stores the number of occurrences of eac word $w$ in each document $d$. Secondly, we group documents by their label and sum up the word's frequency scores. This produces a $P$ x $W$ matrix, where $P$ is the number of unique properties $p$ and $W$ is the number of unique words $w$. 

\begin{table}[htbp]
    \scriptsize
    \centering
    \caption{Example of entries from a Semantic Index. We represent words as a concatenation of their textual form and their part-of-speech. We provide the translation in English between brackets.}

    \begin{tabular}{lrrrrrr}
    \hline
    Word &  occupation &  placeOfBirth &  spouse &  dateOfBirth &  dateOfDeath &  educatedAt \\
    \hline
    né\_VERB (\textit{born})       &       0.002 &         0.327 &   0.007 &        0.731 &        0.599 &       0.000 \\
    27\_NUM         &       0.000 &         0.000 &   0.000 &        0.575 &        0.818 &       0.000 \\
    mort\_VERB (\textit{died})     &       0.000 &         0.004 &   0.000 &        0.000 &        1.000 &       0.000 \\
    % succès\_NOUN    &       0.000 &         1.000 &   0.000 &        0.000 &        0.000 &       0.000 \\
    % peut\_VERB      &       0.307 &         0.863 &   0.275 &        0.000 &        0.000 &       0.293 \\
    % saper\_VERB     &       0.000 &         1.000 &   0.000 &        0.000 &        0.000 &       0.000 \\
    % autorité\_NOUN  &       0.000 &         1.000 &   0.000 &        0.000 &        0.000 &       0.000 \\
    église\_NOUN (\textit{church})   &       0.000 &         1.000 &   0.000 &        0.000 &        0.000 &       0.000 \\
    entamé\_VERB (\textit{started})   &       0.000 &         0.000 &   1.000 &        0.000 &        0.000 &       0.000 \\
    % procédure\_NOUN &       0.000 &         0.000 &   1.000 &        0.000 &        0.000 &       0.000 \\
    ...    &       ... &        ... &   ... &        ... &        ... &       ... \\

    \hline
    \end{tabular}
    \label{tab:semanticindex}
\end{table}

\begin{table}[htbp]
    \centering
    \caption{Distribution of unique words in the Weak and GT Semantic Indices }
    \begin{tabular}{lrr}
    \hline
                            &   Weak Semantic Index     &   GT Semantic Index   \\
        \hline
         Unique words      &   1452                     &   495                   \\ \hline
    \end{tabular}
    \label{tab:semindexstatistic}
\end{table}

\subsection{Extracting and Classifying Relations with the Indices}
\label{sec:indexclassification}

\begin{figure}[htpb]
    \centering
    \includegraphics[width=\textwidth]{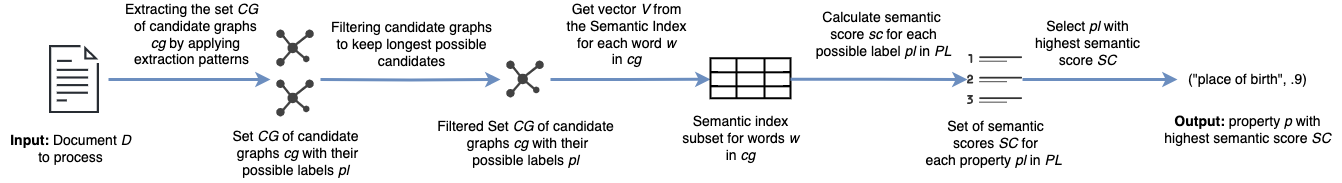}
    \caption{Classification pipeline using the Syntactic and Semantic Indices}
    \label{fig:classificationppipeline}
\end{figure}

When processing documents, we use the Syntactic Index to extract candidate graphs before classifying them with the Semantic Index. This extraction and classification pipeline, which is shown in Figure \ref{fig:classificationppipeline}, is described below.

The first step of the pipeline consists in extracting facts from the dependency graph corresponding to a sentence. To do so, we first search for a corresponding entry in the Syntactic Index for each node in the dependency graph, i.e. any word in the sentence. Then, for each selected entry, we apply the morpho-syntactic extraction patterns to extract candidate subgraphs from the dependency graph. Finally, we filter these candidate subgraphs by only keeping the longest possible sequences. This step outputs the set $CG$ of candidate subgraphs $cg$, each associated with their possible labels $pl$. 

The second step the pipeline consists in classifying each candidate subgraph $cg$. To do so, we first create a $W \times P$ matrix where $W$ is the number of words $w$ from $CG$ found in the Semantic Index, and $P$ is the number of unique properties $p$ in $pl$. From this matrix, we calculate the semantic score $sc$ of each property, i.e. we calculate the harmonic mean of each $p$. This returns a vector $V$ of dimension $P$ containing the semantic score of each $p$. The property $p$ with the highest semantic score $sc$ in $V$ is selected if $sc$ is greater than a semantic threshold $st$. The model outputs "Other" if any of the preceding conditions is not met.

% First, we identify the anchor node of $CG$ and take the set $PA$ of extraction patterns $pa$ for this anchor in the Syntactic Index. Then, we apply the isomorphism test between $CG$ and each pattern $pa$ of the same size as $CG$ to find a matching pattern $mp$. If there is a match, we take the possible labels $pl$ associated with $mp$. 

\section{Evaluation}
\label{sec:evaluation}

In this section, we evaluate the capacity of the Semantic Indices to classify candidate graphs. We do not evaluate the Syntactic Indices as extractor. In order to simulate the extraction step, we first apply the Syntactic Index to each sample in the Development and Test sets to determine what are its possible labels, and provide them to the Semantic Index classifiers.

We evaluate the Weak and GT Semantic Index classifiers in terms of Precision, Recall and F1 scores. Furthermore, we test each classifier on the Development and Test datasets with a semantic threshold between 0 and 0.9. Table \ref{tab:modelevaluation} shows the scores obtained by the Weak and GT Semantic Indices on the Development and Tests datasets for each semantic threshold.

%Thus for each model, we run 40 evaluations, amounting to a total of 80 evaluations.  

\begin{table}[htbp]
    \centering
    \caption{Precision, Recall and F1 scores obtained by the Weak and GT Semantic Indices on the Development and Test datasets for each threshold}

    \begin{tabular}{cc|rrr|rrr|rrr|rrr}
    \hline
    {} & {Threshold} & \multicolumn{3}{c|}{Weak Dev. set} & \multicolumn{3}{c|}{GT Dev. set} & \multicolumn{3}{c|}{Weak Test set} & \multicolumn{3}{c}{GT Test set} \\
    {} & {} & {P} & {R} & {F1} & {P} & {R} & {F1} & {P} & {R} & {F1} & {P} & {R} & {F1} \\
    \hline
    \multirow{10}{*}{\rotatebox{90}{\textbf{Weak Semantic Index}}}

& 0.0  &  0.824 &  \textbf{0.272} &  \textbf{0.402} &  0.816 &  \textbf{0.254} & \textbf{ 0.377} &  \textbf{0.771} &  \textbf{0.393} &  \textbf{0.443} &  0.750 &  \textbf{0.311} &  \textbf{0.378} \\
    & 0.1  &  0.824 &  0.265 &  0.394 &  0.816 &  0.252 &  0.375 &  \textbf{0.771} &  0.346 &  0.388 &  \textbf{0.755} &  0.301 &  0.373 \\
    & 0.2  &  0.824 &  0.259 &  0.387 &  0.816 &  0.252 &  0.375 &  \textbf{0.771} &  0.339 &  0.378 &  \textbf{0.755} &  0.295 &  0.365 \\
    & 0.3  &  0.824 &  0.259 &  0.387 &  0.823 &  0.250 &  0.373 &  \textbf{0.771} &  0.339 &  0.378 &  \textbf{0.755} &  0.295 &  0.365 \\
    & 0.4  &  0.824 &  0.235 &  0.353 &  0.819 &  0.239 &  0.356 &  \textbf{0.771} &  0.332 &  0.371 &  \textbf{0.755} &  0.286 &  0.352 \\
& 0.5  &  \textbf{0.828} &  0.222 &  0.336 &  0.818 &  0.235 &  0.351 &  \textbf{0.771} &  0.318 &  0.354 &  \textbf{0.755} &  0.286 &  0.352 \\
    & 0.6  &  0.687 &  0.204 &  0.307 &  0.785 &  0.216 &  0.320 &  \textbf{0.771} &  0.290 &  0.316 &  0.612 &  0.248 &  0.296 \\
    & 0.7  &  0.687 &  0.188 &  0.290 &  0.784 &  0.209 &  0.312 &  \textbf{0.771} &  0.283 &  0.304 &  0.612 &  0.248 &  0.296 \\
& 0.8  &  0.687 &  0.177 &  0.274 &  \textbf{0.831} &  0.197 &  0.297 &  0.629 &  0.268 &  0.282 &  0.612 &  0.248 &  0.296 \\
    & 0.9  &  0.554 &  0.146 &  0.227 &  0.691 &  0.162 &  0.242 &  0.486 &  0.232 &  0.224 &  0.476 &  0.201 &  0.232 \\

    \hline 

    \multirow{10}{*}{\rotatebox{90}{\textbf{GT Semantic Index}}}

    & 0.0  &  0.773 &  \textbf{0.133} &  \textbf{0.225} &  0.789 &  \textbf{0.109} &  \textbf{0.183} &  \textbf{0.679} &  \textbf{0.227} &  \textbf{0.331} &  0.643 &  \textbf{0.217} &  \textbf{0.313} \\
    & 0.1  &  0.773 &  0.130 &  0.220 &  0.789 &  \textbf{0.109} &  \textbf{0.183} &  0.536 &  0.179 &  0.260 &  0.643 &  0.207 &  0.298 \\
    & 0.2  &  0.773 &  0.130 &  0.220 &  0.789 &  \textbf{0.109} &  \textbf{0.183} &  0.536 &  0.179 &  0.260 &  0.643 &  0.207 &  0.298 \\
    & 0.3  &  0.773 &  0.124 &  0.212 &  0.789 &  0.107 &  0.180 &  0.536 &  0.179 &  0.260 &  0.643 &  0.207 &  0.298 \\
    & 0.4  &  \textbf{0.857} &  0.124 &  0.212 &  \textbf{0.798} &  0.107 &  0.181 &  0.536 &  0.177 &  0.257 &  0.643 &  0.204 &  0.296 \\
    & 0.5  &  \textbf{0.857} &  0.120 &  0.207 &  0.796 &  0.103 &  0.175 &  0.571 &  0.177 &  0.263 &  \textbf{0.714} &  0.195 &  0.294 \\
    & 0.6  &  \textbf{0.857} &  0.116 &  0.201 &  0.795 &  0.097 &  0.166 &  0.571 &  0.177 &  0.263 &  \textbf{0.714} &  0.185 &  0.278 \\
    & 0.7  &  \textbf{0.857} &  0.084 &  0.151 &  0.766 &  0.078 &  0.134 &  0.571 &  0.145 &  0.220 &  \textbf{0.714} &  0.147 &  0.221 \\
    & 0.8  &  \textbf{0.857} &  0.077 &  0.137 &  0.742 &  0.076 &  0.131 &  0.571 &  0.138 &  0.209 &  \textbf{0.714} &  0.147 &  0.221 \\
    & 0.9  &  0.429 &  0.039 &  0.069 &  0.528 &  0.051 &  0.084 &  0.143 &  0.058 &  0.082 &  0.429 &  0.096 &  0.136 \\
    \hline

    \end{tabular}

    \label{tab:modelevaluation}
    \normalsize
\end{table}

On every dataset, the Weak Semantic Index reaches high to very high Precision scores in between 0.75 to 0.83. However, it reaches lower to average Recall scores in between 0.14 to 0.39. Similarly, it reaches F1 scores in between 0.22 to 0.44. The GT Semantic Index also reaches high to very high Precision scores in between 0.74 and 0.85 on the Development dataset. However, its scores are lower on the Test sets and are set in between 0.53 and 0.71. Recall and F1 scores also go from low to average and are set in between 0.3 and 0.22 and 0.6 and 0.31 respectively. Those scores are thus lower than those obtained by the Weak Semantic Index. The lower Recall scores obtained by the GT Semantic Index can be explained by the smaller training set, which limits the amount of syntactic patterns the model can learn. Moreover, we notice every score significantly drops when the semantic threshold is set to 0.9. When aiming for the highest Precision, this suggests the optimal value for the threshold is around 0.6, whereas when aiming for the highest Recall, the optimal value for the threshold is 0. 

As shown in Table \ref{tab:ruleerrors}, most errors are due to either missing anchors or missing patterns in the Syntactic Index. In such cases, the classifier outputs the label "Other". Thus, most errors can be solved by increasing the diversity of morpho-syntactic patterns stored in the Syntactic Index. Other errors are misclassifications by the classifier. Taking contextual information outside of the candidate graph might help in refining the prediction. Extraction patterns may also have been associated with a wrong label because of the distant supervision method. Thus, controlling the output of the annotation step may also improve the results of the classification step.

\begin{table}[htbp]
    \centering
    \caption{Distribution of samples wrongly labelled by the Semantic Index Classifiers on the Development and Test set}

        \begin{tabular}{lrr}
        \hline
        Error type & Weak Semantic Index & GT Semantic Index \\
        \hline
        % Missing anchor or pattern from Syntactic Index & 778 & 924 \\
        % Wrong label given by the classifier & 25 & 16 \\
        Anchor missing from Syntactic Index & 527 & 725 \\
        No matching pattern in the Syntactic Index & 251 & 199 \\
        Wrong label given by the classifier & 25 & 16 \\
        \hline

        \end{tabular}
        \label{tab:ruleerrors}
\end{table}

\section{Conclusion}
\label{sec:conclusion}

In this paper, we have presented a method for the extraction and categorisation of an unrestricted set of relationships from text. Our method relies on high Precision morpho-syntactic extraction patterns collected from Wikipedia and Wikidata by a distant supervision method, and creates Syntactic and Semantic Indices to extract and classify candidate graphs. The evaluations on the six datasets we have built show that our approach can achieve very high Precision scores of up to 0.85, but with lower Recall and F1 scores. Most errors can be corrected by increasing the variety of morpho-syntactic extraction patterns stored in the Syntactic Index. Pre-processing documents with NER may also improve the precision of the patterns. 

Our approach allows to quickly create rule-based systems for Information Extraction and to build large weakly annotated datasets to train machine-learning and deep-learning based classifiers. In future works, we plan to evaluate the performances of the Syntactic Index as an information extractor, and compare the performances of the Semantic Index classifier with machine-learning based classifiers such as Random Forest or XGBoost. Moreover, we plan to evaluate the performances of our model to extract data from documents of different origins, writing styles and languages. 

% we plan to evaluate the impact of changing the hyperparamaters of the Syntactic and Semantic Indices on their performances.

% study how these patterns may be used for joined Named Entity Recognition and Relation Extraction

% \subsubsection{Acknowledgements}
% This research is supported by the Région Bourgogne Franche-Comté, France, as part of the EMONTAL project (Extraction and Ontology Modeling of Subjects and Places for the Exploitation of the Documentary Funds of Bourgogne Franche-Comté, 2020–2023).

\bibliographystyle{splncs04}
\bibliography{bibliography}

\begin{thebibliography}{10}
\providecommand{\url}[1]{\texttt{#1}}
\providecommand{\urlprefix}{URL }
\providecommand{\doi}[1]{https://doi.org/#1}

\bibitem{10.1145/336597.336644}
Agichtein, E., Gravano, L.: Snowball: Extracting relations from large plain-text collections. In: Proceedings of the Fifth ACM Conference on Digital Libraries. p. 85–94. DL '00, Association for Computing Machinery, New York, NY, USA (2000). \doi{10.1145/336597.336644}

\bibitem{akbik-loser-2012-kraken}
Akbik, A., L{\"o}ser, A.: {K}rake{N}: N-ary facts in open information extraction. In: Proceedings of the Joint Workshop on Automatic Knowledge Base Construction and Web-scale Knowledge Extraction ({AKBC}-{WEKEX}). pp. 52--56. Association for Computational Linguistics, Montr{\'e}al, Canada (Jun 2012), \url{https://aclanthology.org/W12-3010}

\bibitem{alfonseca-etal-2012-pattern}
Alfonseca, E., Filippova, K., Delort, J.Y., Garrido, G.: Pattern learning for relation extraction with a hierarchical topic model. In: Proceedings of the 50th Annual Meeting of the Association for Computational Linguistics (Volume 2: Short Papers). pp. 54--59. Association for Computational Linguistics, Jeju Island, Korea (Jul 2012), \url{https://aclanthology.org/P12-2011}

\bibitem{ali2021unsupervised}
Ali, M., Saleem, M., Ngomo, A.C.N.: Unsupervised relation extraction using sentence encoding. In: The Semantic Web: ESWC 2021 Satellite Events: Virtual Event, June 6--10, 2021, Revised Selected Papers 18. pp. 136--140. Springer (2021)

\bibitem{10.1145/2611040.2611079}
Arnold, P., Rahm, E.: Extracting semantic concept relations from wikipedia. In: Proceedings of the 4th International Conference on Web Intelligence, Mining and Semantics (WIMS14). WIMS '14, Association for Computing Machinery, New York, NY, USA (2014). \doi{10.1145/2611040.2611079}

\bibitem{augenstein-etal-2017-semeval}
Augenstein, I., Das, M., Riedel, S., Vikraman, L., McCallum, A.: {S}em{E}val 2017 task 10: {S}cience{IE} - extracting keyphrases and relations from scientific publications. In: Proceedings of the 11th International Workshop on Semantic Evaluation ({S}em{E}val-2017). pp. 546--555. Association for Computational Linguistics, Vancouver, Canada (Aug 2017). \doi{10.18653/v1/S17-2091}

\bibitem{Bekoulis_2018}
Bekoulis, G., Deleu, J., Demeester, T., Develder, C.: Joint entity recognition and relation extraction as a multi-head selection problem. Expert Systems with Applications  \textbf{114},  34--45 (dec 2018). \doi{10.1016/j.eswa.2018.07.032}

\bibitem{bhutani-etal-2016-nested}
Bhutani, N., Jagadish, H.V., Radev, D.: Nested propositions in open information extraction. In: Proceedings of the 2016 Conference on Empirical Methods in Natural Language Processing. pp. 55--64. Association for Computational Linguistics, Austin, Texas (Nov 2016). \doi{10.18653/v1/D16-1006}

\bibitem{Brin1998ExtractingPA}
Brin, S.: Extracting patterns and relations from the world wide web. In: International Workshop on the Web and Databases (1998)

\bibitem{bunescu-mooney-2005-shortest}
Bunescu, R., Mooney, R.: A shortest path dependency kernel for relation extraction. In: Proceedings of Human Language Technology Conference and Conference on Empirical Methods in Natural Language Processing. pp. 724--731. Association for Computational Linguistics, Vancouver, British Columbia, Canada (Oct 2005), \url{https://aclanthology.org/H05-1091}

\bibitem{10.5555/2976248.2976270}
Bunescu, R.C., Mooney, R.J.: Subsequence kernels for relation extraction. In: Proceedings of the 18th International Conference on Neural Information Processing Systems. p. 171–178. NIPS'05, MIT Press, Cambridge, MA, USA (2005)

\bibitem{chan-roth-2011-exploiting}
Chan, Y.S., Roth, D.: Exploiting syntactico-semantic structures for relation extraction. In: Proceedings of the 49th Annual Meeting of the Association for Computational Linguistics: Human Language Technologies. pp. 551--560. Association for Computational Linguistics, Portland, Oregon, USA (Jun 2011), \url{https://aclanthology.org/P11-1056}

\bibitem{chen-etal-2020-joint-entity}
Chen, Y., Sun, Y., Yang, Z., Lin, H.: Joint entity and relation extraction for legal documents with legal feature enhancement. In: Proceedings of the 28th International Conference on Computational Linguistics. pp. 1561--1571. International Committee on Computational Linguistics, Barcelona, Spain (Online) (Dec 2020). \doi{10.18653/v1/2020.coling-main.137}

\bibitem{christopoulou-etal-2018-walk}
Christopoulou, F., Miwa, M., Ananiadou, S.: A walk-based model on entity graphs for relation extraction. In: Proceedings of the 56th Annual Meeting of the Association for Computational Linguistics (Volume 2: Short Papers). pp. 81--88. Association for Computational Linguistics, Melbourne, Australia (Jul 2018). \doi{10.18653/v1/P18-2014}

\bibitem{collins2001convolution}
Collins, M., Duffy, N.: Convolution kernels for natural language. Advances in neural information processing systems  \textbf{14} (2001)

\bibitem{cui2018neural}
Cui, L., Wei, F., Zhou, M.: Neural open information extraction (2018)

\bibitem{del2013clausie}
Del~Corro, L., Gemulla, R.: Clausie: clause-based open information extraction. In: Proceedings of the 22nd international conference on World Wide Web. pp. 355--366 (2013)

\bibitem{devlin-etal-2019-bert}
Devlin, J., Chang, M.W., Lee, K., Toutanova, K.: {BERT}: Pre-training of deep bidirectional transformers for language understanding. In: Proceedings of the 2019 Conference of the North {A}merican Chapter of the Association for Computational Linguistics: Human Language Technologies, Volume 1 (Long and Short Papers). pp. 4171--4186. Association for Computational Linguistics, Minneapolis, Minnesota (Jun 2019). \doi{10.18653/v1/N19-1423}

\bibitem{doddington-etal-2004-automatic}
Doddington, G., Mitchell, A., Przybocki, M., Ramshaw, L., Strassel, S., Weischedel, R.: The automatic content extraction ({ACE}) program {--} tasks, data, and evaluation. In: Proceedings of the Fourth International Conference on Language Resources and Evaluation ({LREC}{'}04). European Language Resources Association (ELRA), Lisbon, Portugal (May 2004), \url{http://www.lrec-conf.org/proceedings/lrec2004/pdf/5.pdf}

\bibitem{Elsahar_2017}
Elsahar, H., Demidova, E., Gottschalk, S., Gravier, C., Laforest, F.: Unsupervised open relation extraction. In: Lecture Notes in Computer Science, pp. 12--16. Springer International Publishing (2017). \doi{10.1007/978-3-319-70407-4_3}, \url{https://doi.org/10.1007\%2F978-3-319-70407-4_3}

\bibitem{fader2011identifying}
Fader, A., Soderland, S., Etzioni, O.: Identifying relations for open information extraction. In: Proceedings of the 2011 conference on empirical methods in natural language processing. pp. 1535--1545 (2011)

\bibitem{gabor-etal-2018-semeval}
G{\'a}bor, K., Buscaldi, D., Schumann, A.K., QasemiZadeh, B., Zargayouna, H., Charnois, T.: {S}em{E}val-2018 task 7: Semantic relation extraction and classification in scientific papers. In: Proceedings of the 12th International Workshop on Semantic Evaluation. pp. 679--688. Association for Computational Linguistics, New Orleans, Louisiana (Jun 2018). \doi{10.18653/v1/S18-1111}

\bibitem{Gabrilovich2007ComputingSR}
Gabrilovich, E., Markovitch, S.: Computing semantic relatedness using wikipedia-based explicit semantic analysis. In: International Joint Conference on Artificial Intelligence (2007), \url{https://api.semanticscholar.org/CorpusID:5291693}

\bibitem{girju-etal-2007-semeval}
Girju, R., Nakov, P., Nastase, V., Szpakowicz, S., Turney, P., Yuret, D.: {S}em{E}val-2007 task 04: Classification of semantic relations between nominals. In: Proceedings of the Fourth International Workshop on Semantic Evaluations ({S}em{E}val-2007). pp. 13--18. Association for Computational Linguistics, Prague, Czech Republic (Jun 2007), \url{https://aclanthology.org/S07-1003}

\bibitem{han2020data}
Han, X., Gao, T., Lin, Y., Peng, H., Yang, Y., Xiao, C., Liu, Z., Li, P., Sun, M., Zhou, J.: More data, more relations, more context and more openness: A review and outlook for relation extraction (2020)

\bibitem{https://doi.org/10.48550/arxiv.1810.10147}
Han, X., Zhu, H., Yu, P., Wang, Z., Yao, Y., Liu, Z., Sun, M.: Fewrel: A large-scale supervised few-shot relation classification dataset with state-of-the-art evaluation (2018). \doi{10.48550/ARXIV.1810.10147}

\bibitem{han-etal-2018-fewrel}
Han, X., Zhu, H., Yu, P., Wang, Z., Yao, Y., Liu, Z., Sun, M.: {F}ew{R}el: A large-scale supervised few-shot relation classification dataset with state-of-the-art evaluation. In: Proceedings of the 2018 Conference on Empirical Methods in Natural Language Processing. pp. 4803--4809. Association for Computational Linguistics, Brussels, Belgium (Oct-Nov 2018). \doi{10.18653/v1/D18-1514}

\bibitem{hasegawa-etal-2004-discovering}
Hasegawa, T., Sekine, S., Grishman, R.: Discovering relations among named entities from large corpora. In: Proceedings of the 42nd Annual Meeting of the Association for Computational Linguistics ({ACL}-04). pp. 415--422. Barcelona, Spain (Jul 2004). \doi{10.3115/1218955.1219008}

\bibitem{hearst-1992-automatic}
Hearst, M.A.: Automatic acquisition of hyponyms from large text corpora. In: {COLING} 1992 Volume 2: The 14th {I}nternational {C}onference on {C}omputational {L}inguistics (1992), \url{https://aclanthology.org/C92-2082}

\bibitem{hendrickx-etal-2010-semeval}
Hendrickx, I., Kim, S.N., Kozareva, Z., Nakov, P., {\'O}~S{\'e}aghdha, D., Pad{\'o}, S., Pennacchiotti, M., Romano, L., Szpakowicz, S.: {S}em{E}val-2010 task 8: Multi-way classification of semantic relations between pairs of nominals. In: Proceedings of the 5th International Workshop on Semantic Evaluation. pp. 33--38. Association for Computational Linguistics, Uppsala, Sweden (Jul 2010), \url{https://aclanthology.org/S10-1006}

\bibitem{csse.2022.030759}
Huiyu~Sun, R.G.: Lexicalized dependency paths based supervised learning for relation extraction. Computer Systems Science and Engineering  \textbf{43}(3),  861--870 (2022). \doi{10.32604/csse.2022.030759}

\bibitem{jurafsky-martin-speech-2008}
Jurafsky, D., Martin, J.: Speech and Language Processing: An Introduction to Natural Language Processing, Computational Linguistics, and Speech Recognition, vol.~2 (02 2008)

\bibitem{kambhatla-2004-combining}
Kambhatla, N.: Combining lexical, syntactic, and semantic features with maximum entropy models for information extraction. In: Proceedings of the {ACL} Interactive Poster and Demonstration Sessions. pp. 178--181. Association for Computational Linguistics, Barcelona, Spain (Jul 2004), \url{https://aclanthology.org/P04-3022}

\bibitem{kassner-etal-2021-multilingual}
Kassner, N., Dufter, P., Sch{\"u}tze, H.: Multilingual {LAMA}: Investigating knowledge in multilingual pretrained language models. In: Proceedings of the 16th Conference of the European Chapter of the Association for Computational Linguistics: Main Volume. pp. 3250--3258. Association for Computational Linguistics, Online (Apr 2021). \doi{10.18653/v1/2021.eacl-main.284}

\bibitem{lopez-de-lacalle-lapata-2013-unsupervised}
Lopez~de Lacalle, O., Lapata, M.: Unsupervised relation extraction with general domain knowledge. In: Proceedings of the 2013 Conference on Empirical Methods in Natural Language Processing. pp. 415--425. Association for Computational Linguistics, Seattle, Washington, USA (Oct 2013), \url{https://aclanthology.org/D13-1040}

\bibitem{Lin2001DIRTD}
Lin, D., Pantel, P.: Dirt – discovery of inference rules from text (2001)

\bibitem{luan-etal-2018-multi}
Luan, Y., He, L., Ostendorf, M., Hajishirzi, H.: Multi-task identification of entities, relations, and coreference for scientific knowledge graph construction. In: Proceedings of the 2018 Conference on Empirical Methods in Natural Language Processing. pp. 3219--3232. Association for Computational Linguistics, Brussels, Belgium (Oct-Nov 2018). \doi{10.18653/v1/D18-1360}

\bibitem{mausam-etal-2012-open}
{Mausam}, Schmitz, M., Soderland, S., Bart, R., Etzioni, O.: Open language learning for information extraction. In: Proceedings of the 2012 Joint Conference on Empirical Methods in Natural Language Processing and Computational Natural Language Learning. pp. 523--534. Association for Computational Linguistics, Jeju Island, Korea (Jul 2012), \url{https://aclanthology.org/D12-1048}

\bibitem{mintz-etal-2009-distant}
Mintz, M., Bills, S., Snow, R., Jurafsky, D.: Distant supervision for relation extraction without labeled data. In: Proceedings of the Joint Conference of the 47th Annual Meeting of the {ACL} and the 4th International Joint Conference on Natural Language Processing of the {AFNLP}. pp. 1003--1011. Association for Computational Linguistics, Suntec, Singapore (Aug 2009), \url{https://aclanthology.org/P09-1113}

\bibitem{nebhi2013rule}
Nebhi, K.: A rule-based relation extraction system using dbpedia and syntactic parsing. In: Proceedings of the NLP-DBPEDIA-2013 Workshop co-located with the 12th International Semantic Web Conference (ISWC 2013) (2013)

\bibitem{nguyen-etal-2007-subtree}
Nguyen, D.P., Matsuo, Y., Ishizuka, M.: Subtree mining for relation extraction from {W}ikipedia. In: Human Language Technologies 2007: The Conference of the North {A}merican Chapter of the Association for Computational Linguistics; Companion Volume, Short Papers. pp. 125--128. Association for Computational Linguistics, Rochester, New York (Apr 2007), \url{https://aclanthology.org/N07-2032}

\bibitem{niklaus-etal-2018-survey}
Niklaus, C., Cetto, M., Freitas, A., Handschuh, S.: A survey on open information extraction. In: Proceedings of the 27th International Conference on Computational Linguistics. pp. 3866--3878. Association for Computational Linguistics, Santa Fe, New Mexico, USA (Aug 2018), \url{https://aclanthology.org/C18-1326}

\bibitem{pawar2017relation}
Pawar, S., Palshikar, G.K., Bhattacharyya, P.: Relation extraction : A survey (2017)

\bibitem{phi-etal-2018-ranking}
Phi, V.T., Santoso, J., Shimbo, M., Matsumoto, Y.: Ranking-based automatic seed selection and noise reduction for weakly supervised relation extraction. In: Proceedings of the 56th Annual Meeting of the Association for Computational Linguistics (Volume 2: Short Papers). pp. 89--95. Association for Computational Linguistics, Melbourne, Australia (Jul 2018). \doi{10.18653/v1/P18-2015}

\bibitem{popovic-etal-2022-aifb}
Popovic, N., Laurito, W., F{\"a}rber, M.: {AIFB}-{W}eb{S}cience at {S}em{E}val-2022 task 12: Relation extraction first - using relation extraction to identify entities. In: Proceedings of the 16th International Workshop on Semantic Evaluation (SemEval-2022). pp. 1687--1694. Association for Computational Linguistics, Seattle, United States (Jul 2022). \doi{10.18653/v1/2022.semeval-1.232}

\bibitem{quirk-poon-2017-distant}
Quirk, C., Poon, H.: Distant supervision for relation extraction beyond the sentence boundary. In: Proceedings of the 15th Conference of the {E}uropean Chapter of the Association for Computational Linguistics: Volume 1, Long Papers. pp. 1171--1182. Association for Computational Linguistics, Valencia, Spain (Apr 2017), \url{https://aclanthology.org/E17-1110}

\bibitem{Riedel2010ModelingRA}
Riedel, S., Yao, L., McCallum, A.: Modeling relations and their mentions without labeled text. In: ECML/PKDD (2010)

\bibitem{rink-harabagiu-2010-utd}
Rink, B., Harabagiu, S.: {UTD}: Classifying semantic relations by combining lexical and semantic resources. In: Proceedings of the 5th International Workshop on Semantic Evaluation. pp. 256--259. Association for Computational Linguistics, Uppsala, Sweden (Jul 2010), \url{https://aclanthology.org/S10-1057}

\bibitem{ro-etal-2020-multi}
Ro, Y., Lee, Y., Kang, P.: {M}ulti{\^{}}2{OIE}: Multilingual open information extraction based on multi-head attention with {BERT}. In: Findings of the Association for Computational Linguistics: EMNLP 2020. pp. 1107--1117. Association for Computational Linguistics, Online (Nov 2020). \doi{10.18653/v1/2020.findings-emnlp.99}

\bibitem{roth-yih-2004-linear}
Roth, D., Yih, W.t.: A linear programming formulation for global inference in natural language tasks. In: Proceedings of the Eighth Conference on Computational Natural Language Learning ({C}o{NLL}-2004) at {HLT}-{NAACL} 2004. pp.~1--8. Association for Computational Linguistics, Boston, Massachusetts, USA (May 6 - May 7 2004), \url{https://aclanthology.org/W04-2401}

\bibitem{schneider-etal-2017-analysing}
Schneider, R., Oberhauser, T., Klatt, T., Gers, F.A., L{\"o}ser, A.: Analysing errors of open information extraction systems. In: Proceedings of the First Workshop on Building Linguistically Generalizable {NLP} Systems. pp. 11--18. Association for Computational Linguistics, Copenhagen, Denmark (Sep 2017). \doi{10.18653/v1/W17-5402}

\bibitem{seganti-etal-2021-multilingual}
Seganti, A., Firl{\k{a}}g, K., Skowronska, H., Sat{\l}awa, M., Andruszkiewicz, P.: Multilingual entity and relation extraction dataset and model. In: Proceedings of the 16th Conference of the European Chapter of the Association for Computational Linguistics: Main Volume. pp. 1946--1955. Association for Computational Linguistics, Online (Apr 2021). \doi{10.18653/v1/2021.eacl-main.166}

\bibitem{stanovsky-dagan-2016-creating}
Stanovsky, G., Dagan, I.: Creating a large benchmark for open information extraction. In: Proceedings of the 2016 Conference on Empirical Methods in Natural Language Processing. pp. 2300--2305. Association for Computational Linguistics, Austin, Texas (Nov 2016). \doi{10.18653/v1/D16-1252}

\bibitem{stanovsky-etal-2018-supervised}
Stanovsky, G., Michael, J., Zettlemoyer, L., Dagan, I.: Supervised open information extraction. In: Proceedings of the 2018 Conference of the North {A}merican Chapter of the Association for Computational Linguistics: Human Language Technologies, Volume 1 (Long Papers). pp. 885--895. Association for Computational Linguistics, New Orleans, Louisiana (Jun 2018). \doi{10.18653/v1/N18-1081}

\bibitem{swampillai-stevenson-2011-extracting}
Swampillai, K., Stevenson, M.: Extracting relations within and across sentences. In: Proceedings of the International Conference Recent Advances in Natural Language Processing 2011. pp. 25--32. Association for Computational Linguistics, Hissar, Bulgaria (Sep 2011), \url{https://aclanthology.org/R11-1004}

\bibitem{wang2022deep}
Wang, H., Qin, K., Zakari, R.Y., Lu, G., Yin, J.: Deep neural network-based relation extraction: an overview. Neural Computing and Applications pp. 1--21 (2022)

\bibitem{white2016universal}
White, A.S., Reisinger, D., Sakaguchi, K., Vieira, T., Zhang, S., Rudinger, R., Rawlins, K., Van~Durme, B.: Universal decompositional semantics on universal dependencies. In: Proceedings of the 2016 Conference on Empirical Methods in Natural Language Processing. pp. 1713--1723 (2016)

\bibitem{wu-weld-2010-open}
Wu, F., Weld, D.S.: Open information extraction using {W}ikipedia. In: Proceedings of the 48th Annual Meeting of the Association for Computational Linguistics. pp. 118--127. Association for Computational Linguistics, Uppsala, Sweden (Jul 2010), \url{https://aclanthology.org/P10-1013}

\bibitem{yahya-etal-2014-renoun}
Yahya, M., Whang, S., Gupta, R., Halevy, A.: {R}e{N}oun: Fact extraction for nominal attributes. In: Proceedings of the 2014 Conference on Empirical Methods in Natural Language Processing ({EMNLP}). pp. 325--335. Association for Computational Linguistics, Doha, Qatar (Oct 2014). \doi{10.3115/v1/D14-1038}

\bibitem{yan-etal-2009-unsupervised}
Yan, Y., Okazaki, N., Matsuo, Y., Yang, Z., Ishizuka, M.: Unsupervised relation extraction by mining {W}ikipedia texts using information from the web. In: Proceedings of the Joint Conference of the 47th Annual Meeting of the {ACL} and the 4th International Joint Conference on Natural Language Processing of the {AFNLP}. pp. 1021--1029. Association for Computational Linguistics, Suntec, Singapore (Aug 2009), \url{https://aclanthology.org/P09-1115}

\bibitem{yao-etal-2011-structured}
Yao, L., Haghighi, A., Riedel, S., McCallum, A.: Structured relation discovery using generative models. In: Proceedings of the 2011 Conference on Empirical Methods in Natural Language Processing. pp. 1456--1466. Association for Computational Linguistics, Edinburgh, Scotland, UK. (Jul 2011), \url{https://aclanthology.org/D11-1135}

\bibitem{yao-etal-2012-unsupervised}
Yao, L., Riedel, S., McCallum, A.: Unsupervised relation discovery with sense disambiguation. In: Proceedings of the 50th Annual Meeting of the Association for Computational Linguistics (Volume 1: Long Papers). pp. 712--720. Association for Computational Linguistics, Jeju Island, Korea (Jul 2012), \url{https://aclanthology.org/P12-1075}

\bibitem{yao-etal-2019-docred}
Yao, Y., Ye, D., Li, P., Han, X., Lin, Y., Liu, Z., Liu, Z., Huang, L., Zhou, J., Sun, M.: {D}oc{RED}: A large-scale document-level relation extraction dataset. In: Proceedings of the 57th Annual Meeting of the Association for Computational Linguistics. pp. 764--777. Association for Computational Linguistics, Florence, Italy (Jul 2019). \doi{10.18653/v1/P19-1074}

\bibitem{yates-etal-2007-textrunner}
Yates, A., Banko, M., Broadhead, M., Cafarella, M., Etzioni, O., Soderland, S.: {T}ext{R}unner: Open information extraction on the web. In: Proceedings of Human Language Technologies: The Annual Conference of the North {A}merican Chapter of the Association for Computational Linguistics ({NAACL}-{HLT}). pp. 25--26. Association for Computational Linguistics, Rochester, New York, USA (Apr 2007), \url{https://aclanthology.org/N07-4013}

\bibitem{zaporojets2021dwie}
Zaporojets, K., Deleu, J., Develder, C., Demeester, T.: Dwie: an entity-centric dataset for multi-task document-level information extraction (2021)

\bibitem{zeng-etal-2017-incorporating}
Zeng, W., Lin, Y., Liu, Z., Sun, M.: Incorporating relation paths in neural relation extraction. In: Proceedings of the 2017 Conference on Empirical Methods in Natural Language Processing. pp. 1768--1777. Association for Computational Linguistics, Copenhagen, Denmark (Sep 2017). \doi{10.18653/v1/D17-1186}

\bibitem{zhang-etal-2018-graph}
Zhang, Y., Qi, P., Manning, C.D.: Graph convolution over pruned dependency trees improves relation extraction. In: Proceedings of the 2018 Conference on Empirical Methods in Natural Language Processing. pp. 2205--2215. Association for Computational Linguistics, Brussels, Belgium (Oct-Nov 2018). \doi{10.18653/v1/D18-1244}

\bibitem{zhang-etal-2017-position}
Zhang, Y., Zhong, V., Chen, D., Angeli, G., Manning, C.D.: Position-aware attention and supervised data improve slot filling. In: Proceedings of the 2017 Conference on Empirical Methods in Natural Language Processing. pp. 35--45. Association for Computational Linguistics, Copenhagen, Denmark (Sep 2017). \doi{10.18653/v1/D17-1004}

\bibitem{zhila2014open}
Zhila, A., Gelbukh, A.: Open information extraction for spanish language based on syntactic constraints. In: Proceedings of the ACL 2014 Student Research Workshop. pp. 78--85 (2014)

\bibitem{zhou2022survey}
Zhou, S., Yu, B., Sun, A., Long, C., Li, J., Sun, J.: A survey on neural open information extraction: Current status and future directions. arXiv preprint arXiv:2205.11725  (2022)

\bibitem{zhou2020documentlevel}
Zhou, W., Huang, K., Ma, T., Huang, J.: Document-level relation extraction with adaptive thresholding and localized context pooling (2020)

\end{thebibliography}

\end{document}